\documentclass[conference]{IEEEtran}

\usepackage{cite}
\usepackage{amsmath,amssymb,amsfonts}
\usepackage{algorithmic}
\usepackage{graphicx}
\usepackage{textcomp}
\usepackage{xcolor}
\usepackage[colorinlistoftodos]{todonotes}
\def\BibTeX{{\rm B\kern-.05em{\sc i\kern-.025em b}\kern-.08em
    T\kern-.1667em\lower.7ex\hbox{E}\kern-.125emX}}
    
\newcommand{\inx}{\mathrm{in}}
\newcommand{\crossx}{\mathrm{cross}}

\begin{document}

\title{A Framework for Learning Invariant Physical Relations in Multimodal Sensory Processing}

\author{\IEEEauthorblockN{1\textsuperscript{st} Du Xiaorui}
\IEEEauthorblockA{\textit{Fakult\"at Informatik} \\
\textit{Technische Hochschule Ingolstadt}\\
Ingolstadt, Germany\\
xiaoruipython@gmail.com}
\and
\IEEEauthorblockN{2\textsuperscript{nd} Yavuzhan Erdem}
\IEEEauthorblockA{\textit{Fakult\"at Elektro- und Informationstechnik} \\
\textit{Technische Hochschule Ingolstadt}\\
Ingolstadt, Germany\\
erdemyavuzhan2023@gmail.com}
\and
\IEEEauthorblockN{3\textsuperscript{rd} Immanuel Schweizer}
\IEEEauthorblockA{\textit{Artificial Intelligence Research} \\
\textit{Merck KGaA}\\
Darmstadt, Germany\\
immanuel.schweizer@merckgroup.com}
\and
\IEEEauthorblockN{4\textsuperscript{th} Cristian Axenie}
\IEEEauthorblockA{\textit{Audi Konfuzius-Institut Ingolstadt} \\
\textit{Technische Hochschule Ingolstadt}\\
Ingolstadt, Germany \\
cristian.axenie@audi-konfuzius-institut-ingolstadt.de}
}

\maketitle

\begin{abstract}
Perceptual learning enables humans to recognize and represent stimuli invariant to various transformations and build a consistent representation of the self and physical world. Such representations preserve the invariant physical relations among the multiple perceived sensory cues. 

This work is an attempt to exploit these principles in an engineered system. We design a novel neural network architecture capable of learning, in an unsupervised manner, relations among multiple sensory cues. The system combines computational principles, such as competition, cooperation and correlation, in a neurally plausible computational substrate. It achieves that through a parallel and distributed processing architecture in which the relations among the multiple sensory quantities are extracted from time sequenced data.

We describe the core system functionality when learning arbitrary non-linear relations in low-dimensional sensory data. Here, an initial benefit rises from the fact that such a network can be engineered in a relatively straightforward way without prior information about the sensors and their interactions. Moreover, alleviating the need for tedious modelling and parametrization, the network converges to a consistent description of any arbitrary high-dimensional multisensory setup. We demonstrate this through a real-world learning problem, where, from standard RGB camera frames, the network learns the relations between physical quantities such as light intensity, spatial gradient, and optical flow, describing a visual scene.

Overall, the benefits of such a framework lie in the capability to learn non-linear pairwise relations among sensory streams in an architecture that is stable under noise and missing sensor input.

\end{abstract}


\section{Introduction}

Perception is a process of acquiring information that happens over time \cite{adolph2015gibson}. The primary source of perceptual information are events, or "happenings over time" \cite{gibson1969principles}. These events are the critical component of what is kept during perceptual learning and development.
Moreover, multimodal processing emerges in the context of an event \cite{gibson1979ecological}. For example, optic flow and motion parallax, emerge as one moves through the world, whereas accretion and deletion of visual texture elements occur when an object or part of the scene becomes progressively uncovered or occluded as a result of motion \cite{johansson1973visual}.

In any multi-sensory system some aspects of the world and the self will change during a given event. But some aspects will not, they stay invariant under the transformation. These invariances must be extracted from the multi-sensory system to learn more about the world and the self. One such invariance is the physical relations between the different modalities. These relations specify what is permanent, what is changing, and how the change is occurring \cite{gibson1979ecological}. For example, one's motion generates particular transformation patterns in the optic array and it is through these patterns that one identifies unitary objects and simultaneously the trajectory and type of motion through the world. 

Such invariant relations bind different sensory modalities into what is a consistent perception of the scene. But in most engineered systems these relations are specified at design-time. It would be more plausible to believe relations between multiple sensory systems are learned to allow for adaptations should a concept drift or concept shift occur.


Starting from these principles we propose a biologically plausible neural network capable of extracting, in an unsupervised manner, the invariant physical relations among multiple sensory cues. The network employs neural computing principles such as competition, cooperation, and correlation in neural populations, to learn multimodal sensory relations without any prior knowledge about the type of sensory data and the underlying relations. 

We describe the capabilities of such a network in a series of experiments on amodal data (e.g. timeseries) in order to give the reader an understanding over the processing mechanisms of the framework. Subsequently, we demonstrate the framework's applicability in learning invariant physical relations in the visual scene autonomously, without describing the scene mathematically. Once such relations have been learned the system can simultaneously de-noise potentially perturbed sensory streams and even infer missing ones without bringing any modification to the its structure.

\section{Network architecture}

\begin{figure}[!ht]
\centering
\includegraphics [scale=0.4] {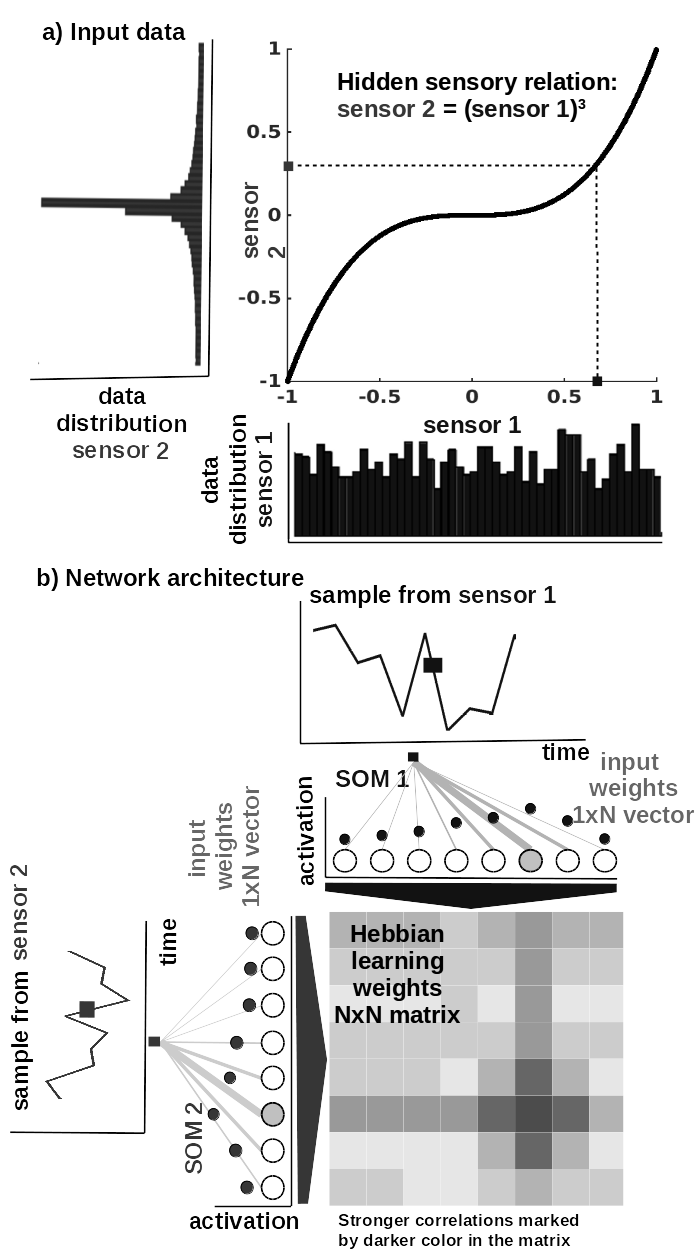}
\caption{Basic functionality. a) Input data resembling a non-linear relation and its distribution - relation is hidden in the data. b) Basic architecture of the network.} 
\label{fig1}
\end{figure}

The adaptive development of the human perceptual capabilities seems to depend strongly on the available sensory inputs, which gradually sharpen their interaction during development, given the constraints imposed by multisensory relations \cite{westermann2007neuroconstructivism}.
Following this principle, we propose a model based on Self-Organizing Maps (SOM) \cite{kohonen1982self} and Hebbian Learning (HL) \cite{chen2008correlative} as main components for learning underlying relations among multiple sensory cues.
In order to introduce the proposed network, we provide a simple example in Figure~\ref{fig1}. 
In this simple example, we consider two input sensor streams following a power law dependency relation (cmp.  Figure~\ref{fig1}a). The system has no prior information about sensory data distributions and the generating processes, but learns the underlying (i.e. hidden) relation directly from the input data in an unsupervised manner.

\subsection{Core model}

The input SOMs extract the distribution of the incoming sensory data, depicted in Figure~\ref{fig1}a, and encode sensory samples in a distributed activity pattern, as shown in Figure~\ref{fig1}b. 
This activity pattern is generated such that the closest preferred value of a neuron to the input sample will be strongly activated and will decay, proportional with distance, for neighbouring units. 
\begin{figure}[!ht]
\centering
\includegraphics [scale=0.4] {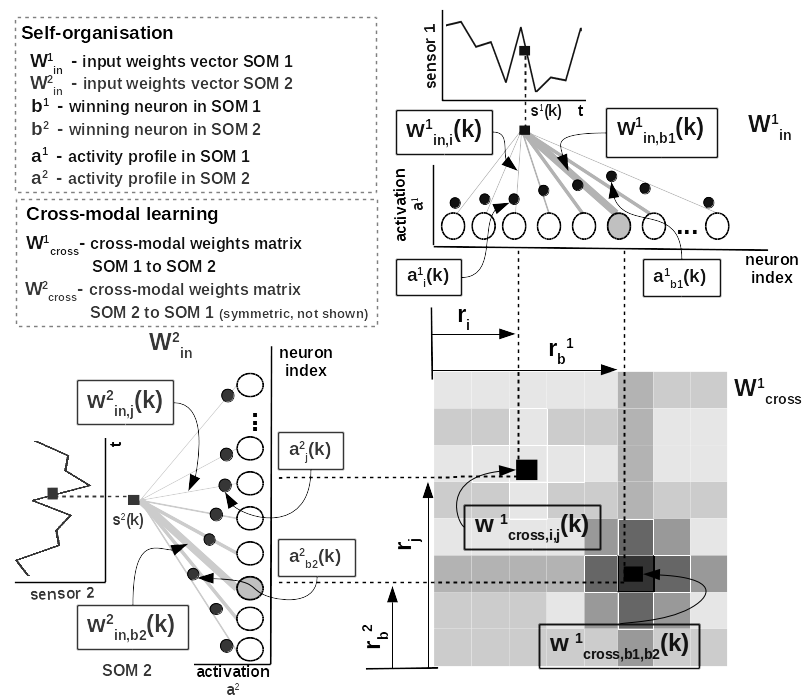}
\caption{Detailed network functionality} 
\label{fig2}
\end{figure}
The SOM specialises to represent a certain (preferred) value in the sensory space and learns its sensitivity (i.e. tuning curve shape).
Given an input sample from the sensor timeseries, $s^p(k)$ at time step $k$, the network follows the processing stages depicted in Figure~\ref{fig2}.
For each $i$-th neuron in the $p$-th input SOM, with preferred value $w_{\inx,i}^{p}$ and tuning curve size ${\xi _{{i}} ^{{p}}(k)}$, the elicited activation is given by
\begin{equation}
a_{i}^{{p}} (k) = \frac {1} {\sqrt{2 \pi} \xi_{{i}}^{{p}}(k)} 
e^{\frac {-(s ^{{p}} (k) - w_{\inx,i} ^{{p}} (k))^2} {2 {\xi _{{i}} ^{{p}}(k)}^2}}.
\label{eq1}
\end{equation}
The winning neuron of the $p$-th population, $ b^{p}(k)$, is the one which elicits the highest activation given the sensory input at time $k$
\begin{equation}
b ^{{p}} (k) = \underset{i}{\mathrm{argmax}} \ {a_i^{{p}} (k)}.
\label{eq2}
\end{equation}
The competition for highest activation in the SOM is followed by cooperation in representing the input sensory space.
Given the winner neuron, $ b^{p}(k)$, the (cooperation) interaction kernel, 
\begin{equation}
h_{b,i} ^{{p}} (k) = e^{\frac {-|| r_{i} -  r_{b} || ^2} {2 {\sigma(k)}^2}}.
\label{eq3}
\end{equation}
allows neighbouring cells (found at position  $r_{i}$ in the network) to precisely represent the sensory input sample given their location in the neighbourhood $\sigma(k)$ of the winning neuron. 
The interaction kernel in Equation~\ref{eq3}, ensures that specific neurons in the network specialise on different areas in the sensory space, such that the input weights (i.e. preferred values) of the neurons are pulled closer to the input sample,
\begin{equation}
\Delta w_{\inx,i}^{{p}} (k) = \alpha(k)h_{b,i}^{{p}} (k)(s^{{p}} (k) - w_{{\inx,i}}^{{p}} (k)).
\label{eq4}
\end{equation}
This corresponds to updating the tuning curves, or sensitivities. Each neuron's tuning curve is modulated by the spatial location of the neuron in the network, the distance to the input sample, the interaction kernel size, and a decaying learning rate $ \alpha(k) $,
\begin{equation}
\Delta \xi_{i}^{{p}} (k) = \alpha(k)h_{b,i}^{{p}} (k)((s^{{p}} (k) - w_{{\inx,i}}^{{p}}(k))^2 - \xi_{i} ^{{p}} (k)^2).
\label{eq5}
\end{equation}
As an illustration of the process, let's consider learned tuning curves shapes for 5 neurons in the input SOMs (i.e. neurons 1, 6, 13, 40, 45), depicted in Figure~\ref{fig3}. We observe that higher input probability distributions are represented by dense and sharp tuning curves (e.g. neuron 1, 6, 13 in SOM1), whereas lower or uniform probability distributions are represented by more sparse and wide tuning curves (e.g. neuron 40, 45 in SOM1).
\begin{figure}[!ht]
\centering
\includegraphics [scale=0.4] {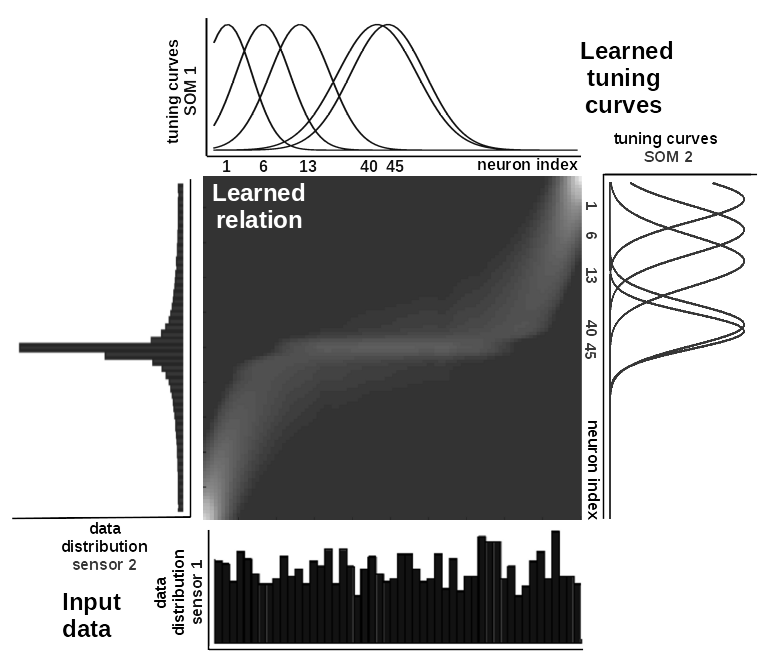}
\caption{Extracted sensory relation and data statistics for the data in Figure~\ref{fig1}a} 
\label{fig3}
\end{figure}
Using this mechanism, the network optimally allocates resources (i.e. neurons). A higher amount of neurons to areas in the input space, which need a finer resolution; and a lower amount for more coarsely represented areas.

Neurons in different SOMs are then linked by a fully (all-to-all) connected matrix of synaptic connections, where the weights in the matrix are computed using Hebbian learning. In the end, the matrix encodes the co-activation patterns between the input layers (i.e. SOMs), as shown in Figure~\ref{fig1}b, and, eventually, the learned relation between the sensory cues, as shown in Figure~\ref{fig3}.
The connections between uncorrelated (or weakly correlated) neurons in each population (i.e. $w_{\crossx}$) are suppressed (i.e. darker color) while correlated neurons connections are enhanced (i.e. brighter color).
The effective correlation pattern encoded in the $w_{\crossx}$ matrix, depicts the actual learnt relation. Mathematically speaking, the physical relation imposes constraints upon possible sensory values - hence supporting de-noising and inference. One can see the network dynamics as a constraint satisfaction problem, whereas the constraint is imposed by the underlying invariant relation among the sensor data (i.e. describing the physics laws).
Formally, the connection weight $w_{\crossx,i,j}^{p}$ between neurons $i, j$ in the different input SOMs are updated with a Hebbian learning rule as follows:
\begin{equation}
\Delta w_{\crossx,i,j}^{{p}} (k) = \eta(k)(a _{i}^{{p}} (k) - \overline{a} _{i}^{{p}}(k))(a _{j} ^{{q}} (k) - \overline{a} _{j} ^{{q}}(k)), 
\label{eq6}
\end{equation}
where
\begin{equation}
\overline{a} _{i} ^{{p}}(k) = (1-\beta(k))\overline{a} _{i} ^{{p}}(k-1) + \beta(k)a _{i}^{{p}}(k),
\label{eq7}
\end{equation}
and $\eta(k)$, $\beta(k)$ are monotonic decaying functions.
Hebbian learning ensures that when neurons fire synchronously their connection strengths increase, whereas if their firing patterns are anti-correlated the weights decrease. 

Self-organisation and Hebbian correlation learning processes evolve simultaneously, such that both the representation and the extracted relation are continuously refined, as new data is available. 

\section{Experiments}

\subsection{Learning in low-dimensional multimodal sensory space}

\begin{figure}[!ht]
\centering
\includegraphics [scale=0.4] {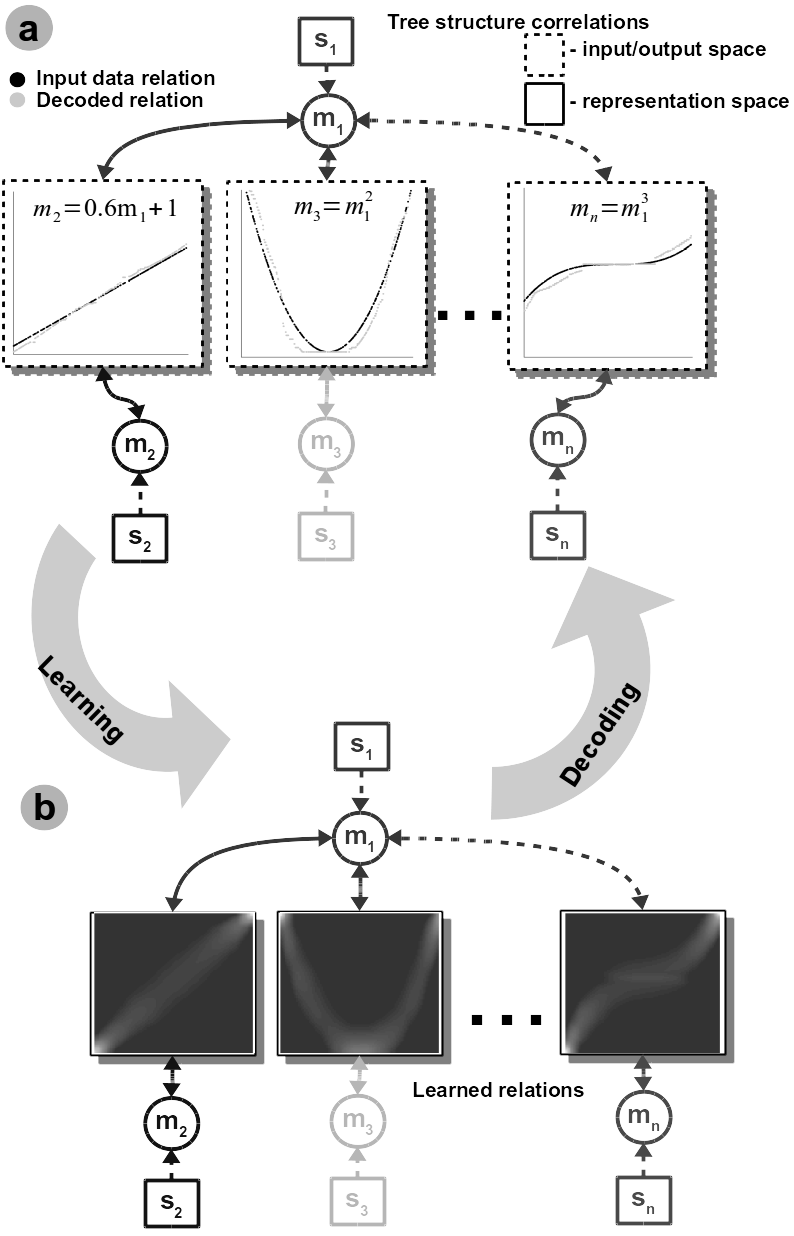}
\caption{Multimodal relations learning in low-dimensional space with the network of relations organized as a tree a) Input data and decoded learnt data. b) Learnt relations among the sensory cues.} 
\label{fig4}
\end{figure}

\begin{figure}[!ht]
\centering
\includegraphics [scale=0.4] {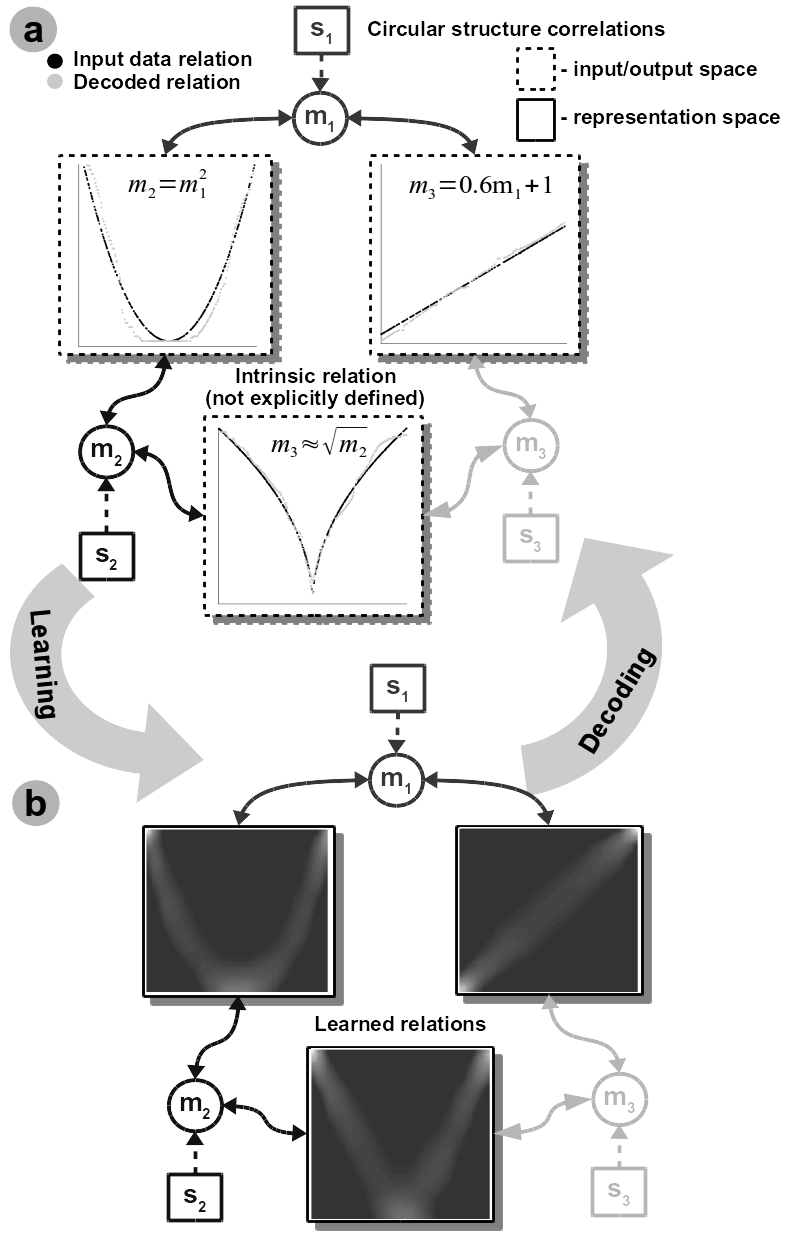}
\caption{Multimodal relations learning in low-dimensional space with circular structure network of relations. a) Input data and decoded learnt data. b) Learnt relations among the sensory cues.} 
\label{fig5}
\end{figure}

In the first batch of experiments we look at how the proposed network can learn arbitrary relations among multiple sensory cues $m_i$ given sensory timeseries $s_i$. For this, we look at two scenarios where the dependencies between sensory cues are organized differently; namely as a tree (Figure~\ref{fig4}) or in a circular structure (Figure~\ref{fig5}). Such architectures impose different local dynamics and a different data flow when extracting the hidden relations.
In both of the cases data from sensory timeseries $s_i$ is fed to the network which encodes each sensory cue $m_i$ in the SOMs and learns the underlying relation in the Hebbian linkage. Using a set of arbitrary relations we demonstrate that the network learns the invariant relations among the interacting sensory cues (Figure~\ref{fig4}b, Figure~\ref{fig5}b), fact validated by the decoded outcome overlaid in the figure over the original (i.e. hidden) relation in the data (Figure~\ref{fig4}a, Figure~\ref{fig5}a).
We use as decoding mechanism an optimisation method that recovers the real-world value given the self-calculated bounds of the input sensory space. The bounds are obtained as minimum and maximum of a cost function of the distance between the current preferred value of the winning neuron and the input sample at the SOM level. Depending on the position of the winning neuron in the $N$-dimensional SOM, the recovered value $y(t)$ is computed as:
\[
 y(t) = 
  \begin{cases} 
   w_{in, i}^p + d_{i}^p & \text{if } i \geq \frac{N}{2}\\
   w_{in, i}^p + d_{i}^p & \text{if } x < \frac{N}{2}
  \end{cases}
\]
where, $ d_{i}^p = \sqrt{2\xi_i^k(k)^2log(\sqrt{2\pi}a_i^p(k)\xi_i^k(k)^2)} $ for the most active neuron with index $i$ in the SOM, a preferred value $w_{in, i}^p$ and $\xi_i^k(k)$ tuning curve size.
The optimiser is based on Brent's method \cite{brent1971algorithm}, a recursive method to find the global optimum of a function for which the analytical form of its derivative is not available or too complex. 

\subsection{Learning in high-dimensional multimodal sensory space}

In the current section we demonstrate that our system is capable of learning invariant sensory relations also in high-dimensional spaces, where the complexity of the interactions goes beyond simple algebrical relations. We considered for our experiments, visual perception where sets of invariant relations within stimulus and self-motion are described through spatio-temporal differentiation and integration  \cite{johansson1976spatio}. For instance, the relation between radius of curvature of the motion path and the two components of acceleration (i.e. tangential and normal) describe consistent visual motion perception through differentiation. 
Such accurate mathematical descriptions, encourage us to hypothesise that a relational model explains visual motion perception based on physical constraints that keep the multiple sensory inputs describing the scene in agreement \cite{johansson1973visual}. Moreover, "information from different modalities belongs together when it is unified by the same invariant relations" \cite{gibson1984development}.

Such premises motivate our second batch of real-world experiments where we demonstrate that the proposed network can learn invariant relations between physical quantities such as light intensity, spatial gradient and optical flow, describing a visual scene from standard RGB camera frames.

Similar work was carried out by \cite{cook2011interacting} for fast event-based (frameless) visual interpretation. Here the authors hard-coded the relations among the sensory cues by deducing simple update formulations from the vectorial representations in the geometry of the scene.
In our experiments we alleviate the need to geometrically describe the problem and the difficulty this analytical formulation poses on the system design. We show how our network learns the underlying relations among the visual scene quantities and, through its dynamics, is able to provide a consistent scene interpretation that brings all sensory quantities in agreement.

In our real-world instantiation the network learns the relations between multiple visual cues, such as: optic flow $F$, light intensity $I$, spatial light intensity gradient $G$, and temporal light intensity derivative $V$. The physics of the visual scene already imposes that $G$ should be the gradient of $I$, and that spatial variation $G$ in brightness should match time variation $V$ according to the local optic flow $F$. Such constraints obey three-dimensional geometry and describe the mathematics of the scene \cite{johansson1976spatio}.
\begin{figure}[!ht]
\centering
\includegraphics [scale=0.30] {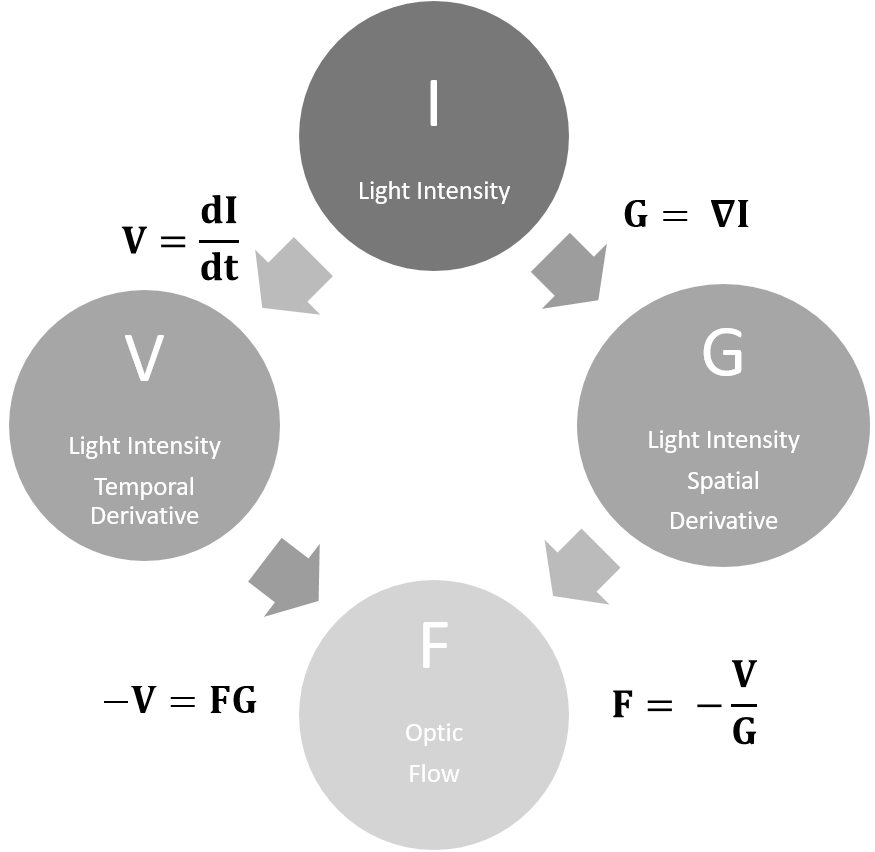}
\caption{Network structure for visual scene interpretation. Physical relations between Optical Flow $F$, Light Intensity $I$, Light Intensity Spatial Gradient $G$ and Intensity Temporal Derivative $V$.} 
\label{fig6}
\end{figure}

For instance, the network describing the relation among $V$, $F$, and $G$, (Figure~\ref{fig6}), namely, $ -V = FG $ shows that the change in brightness over time is given by the speed of the optic flow times the change in brightness in the direction of the optic flow (i.e. optical flow constraint equation \cite{horn1981determining}). Similarly, the relation between $I$ and $G$ requires that $ G = \nabla I $ holds. Yet, the equation $ -V = FG $ can not be solved uniquely for $F$ and $G$, due to the aperture problem. Given a spatial gradient $G$ and a temporal gradient $V$, the solutions for $F$ lie on a vector perpendicular on $G$ (i.e. for a limited aperture size and an edge structure, motion can only be estimated normal to that edge). This theoretical treatment is meant to support the reader to understand the output of the network, which, again, had no prior knowledge about the relations or the sensory quantities fed to it.

In our experiments we fed the network, subsequently with pairs of: light intensity ($I$) and light intensity derivative ($V$); light intensity ($I$) and light intensity spatial gradient ($G$); and light intensity temporal derivative ($V$), spatial gradient ($G$) and optical flow ($F$), respectively. The network learns the individual relations among these quantities from the incoming RGB camera frames (i.e. resolution: 200x200px, color scale: gray(1 channel), framerate: 30FPS). Note that in order to generate all quantities for learning the relations we used existing operators in state-of-the-art libraries (e.g. Sobel operator, Lucas-Kanade optic flow method in OpenCV). After the network learned the individual relations among pairs of sensory streams we connected all of the SOMs corresponding to the sensors and their respective Hebbian linkage.

\begin{figure}[!ht]
\centering
\includegraphics [scale=0.28] {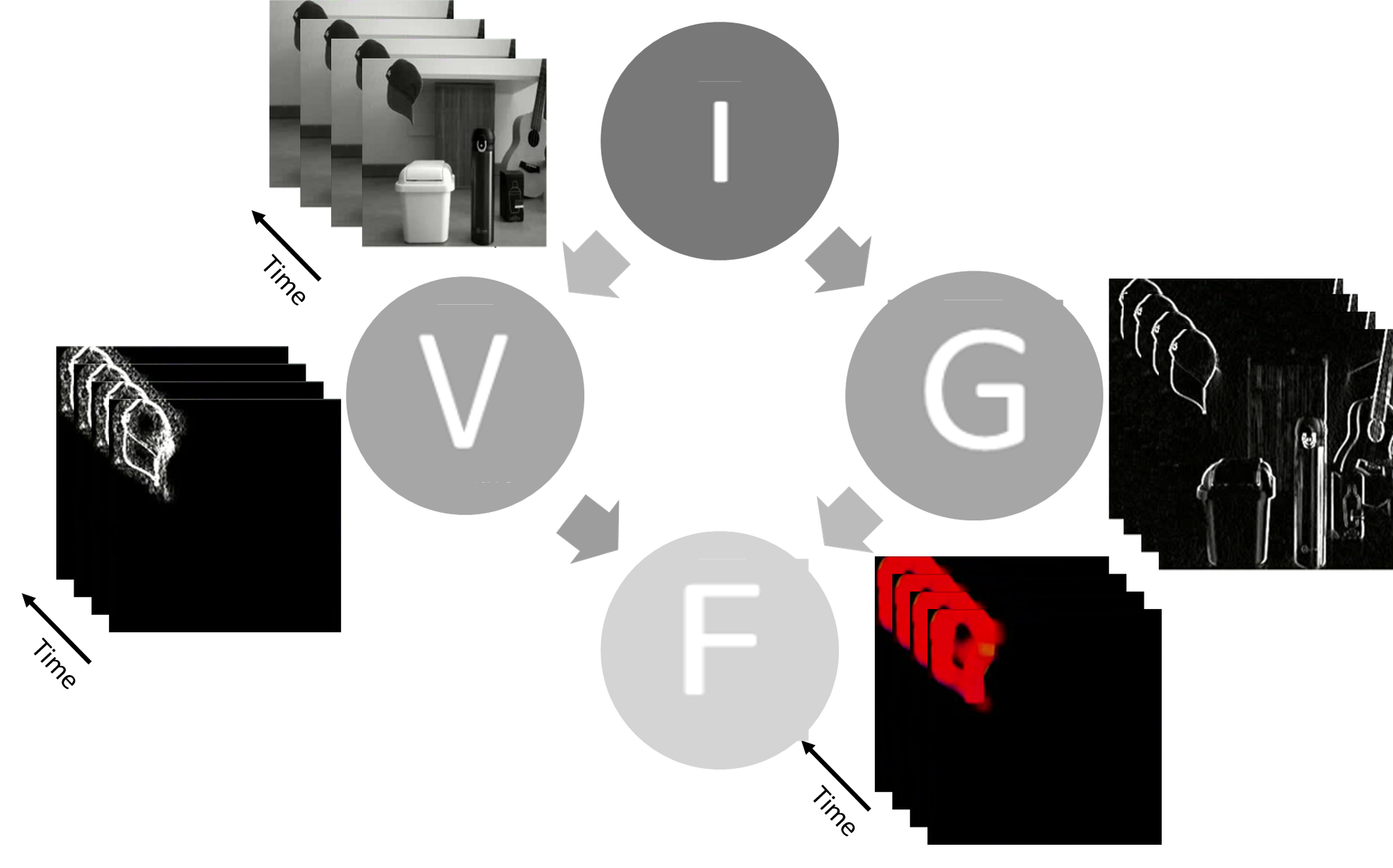}
\caption{Inferred quantities given the learned relations among multimodal cues in visual scene interpretation.} 
\label{fig7}
\end{figure}

The network converges to a stable representation, as shown in Figure~\ref{fig7}. Note that the relations are learned in pairwise manner, subsequently allowing the network to converge given all relations. The network allows for such a decoupling, as shown also in the tree and circular experiments in the previous section. 

\begin{figure}[!ht]
\centering
\includegraphics [scale=0.28] {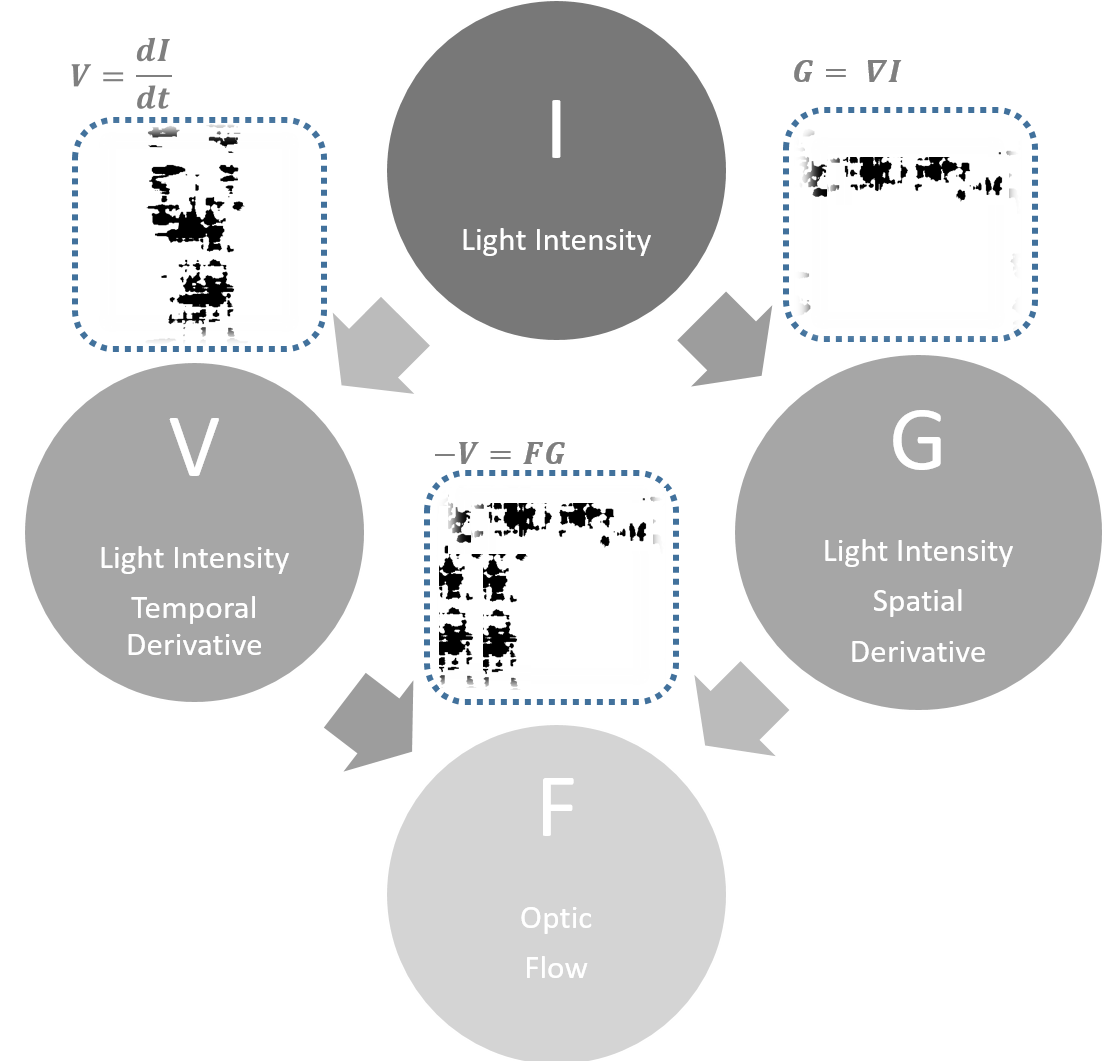}
\caption{Hidden relations among the multimodal sensory cues in visual scene interpretation.} 
\label{fig8}
\end{figure}

The network learns the hidden relations among the multiple modalities describing the visual scene, as shown in Figure~\ref{fig8}.
 
\section{Related work}

In order to frame our contribution we describe comparatively other state-of-the-art approaches addressing the extraction of (unknown) relations among multiple data streams. We briefly address aspects such as the amount of prior information used in the model and the inference capabilities of the models.

Related work in Cook et al.\cite{cook2010unsupervised}, used a combination of simple biologically plausible mechanisms, like Winner-Take-All (WTA) circuitry, Hebbian learning, and homeostatic activity regulation, to extract relations in artificially generated sensory data. After learning, the model was able to infer missing quantities given the learned relations and available sensors, clean-up noisy sensory inputs and integrate the sensory data consistently.

Using a different neurally inspired substrate, Weber et al. \cite{weber2007self} combined competition and cooperation in a self-organizing network of processing units to extract coordinate transformations in a robotic visual object localization scenario. 

Going away from biological inspiration, Mandal et al.\cite{mandal2013non} used a nonlinear canonical correlation analysis method, termed alpha-beta divergence correlation analysis (ABCA),
to extract relations between sets of multidimensional random variables. 
The model employed a probabilistic method based on nonlinear correlation analysis using a more flexible metric (i.e. divergence / distance) than typical canonical correlation analysis.

Using a neurally inspired computing substrate for implementing canonical correlation
analysis Hsieh et al. \cite{hsieh2000nonlinear} proposed a model able to extract the underlying structures between two sets of variables under moderate noise conditions, basically employing nonlinear PCA. 

\textbf{Priors}

Although less intuitive, the purely mathematical approaches \cite{mandal2013non} (i.e. using canonical correlation analysis) need less tuning effort as the parameters are the result of an optimisation procedure.
On the other side, the neurally inspired approaches \cite{cook2010unsupervised, weber2007self} or the hybrid approaches \cite{hsieh2000nonlinear} (i.e. combining neural networks and correlation analysis) need a more judicious parameter tuning, as their dynamics are more sensitive, and can either reach instability (e.g. recurrent networks) or local minima.
Excepting parametrisation, prior information about inputs is generally needed when instantiating the state-of-the-art systems for a certain scenario. 
Sensory values bounds and probability distributions must be explicitly encoded in the models through explicit tiling of tuning values over a population of neurons \cite{cook2010unsupervised, weber2007self}, linear coefficients in vector combinations \cite{mandal2013non}, or standardisation routines of input variables \cite{hsieh2000nonlinear}. Our model doesn't require any specification of the bounds or other information about the input sensory data. It learns autonomously the statistics of the data required for learning (i.e. probability density).

\textbf{Inference}

We consider here the capability to infer (i.e. predict) missing quantities once the hidden relation among sensory quantities is learned.
The capability to use the learned functional relations to determine missing quantities is not available in all presented models, e.g., \cite{mandal2013non} due to the fact that the divergence and correlation coefficient expressions might be non-invertible functions.
On the other side, using either the learned co-activation weight matrix \cite{cook2010unsupervised, weber2007self}, or the known standard deviations of the canonical variants \cite{hsieh2000nonlinear} these models are able to predict missing quantities to some extent. 
Our model is able, once it has learnt the hidden relation, to use it to infer possible values of a sensor given the other available ones. When addressing non-invertible relations (e.g. powerlaw) our system uses the learnt domain and range (i.e. tuning curves and learnt data distribution) to identify the correct inverse function. As the domain and range of the inverse relation come from the range, and domain of the hidden relation, respectively, the system can perform the swapping of domain and range to converge to a correct solution.

\section{Conclusion}

Looking at perception as a hierarchy of physical relations to fulfil is an interesting biological hypothesis. In this work we demonstrated that an engineered system, using principles of neural computation, is able to learn without supervision such relations describing low-dimensional sensory streams as well as high-dimensional visual scenes. The network has the potential to learn from arbitrary sensory streams and provide an interpretation of the multisensory scene without the need to model either the sensors or their interactions. We believe that such a framework can leverage multisensory processing by exploiting underlying relations in real-world sensory streams and their interactions employing simple computational principles towards efficient implementations across various scenarios.

\bibliographystyle{IEEEtran} 

\bibliography{irena-manuscript}

\end{document}